\documentclass[a4paper,fleqn]{cas-sc}

\usepackage[authoryear,longnamesfirst]{natbib}
\usepackage{color,array}
\usepackage{graphicx}
\usepackage{amsmath,amsfonts}
\usepackage{algorithm}
\usepackage{array}
\usepackage[caption=false,font=normalsize,labelfont=sf,textfont=sf]{subfig}
\usepackage{textcomp}
\usepackage{stfloats}
\usepackage{url}
\usepackage{verbatim}
\usepackage[utf8]{inputenc}
\usepackage{bm}
\usepackage{amsmath}
\newcommand\norm[1]{\left\lVert#1\right\rVert}
\usepackage{algpseudocode}
\usepackage{mathtools}
\usepackage{amssymb}
\usepackage{multirow}
\usepackage{array,multirow}
\usepackage{rotating}
\usepackage[authoryear,longnamesfirst]{natbib}

\usepackage{amsthm}
\usepackage{booktabs}
\usepackage{comment}
\usepackage{adjustbox}
  \newtheoremstyle{dotless}{}{}{\itshape}{}{\bfseries}{}{ }{}
  \theoremstyle{dotless}
  \newtheorem{thm}{Theorem}
  \newtheorem{prop}{Proposition}

\def\tsc#1{\csdef{#1}{\textsc{\lowercase{#1}}\xspace}}
\tsc{WGM}
\tsc{QE}

\begin{document}
\let\WriteBookmarks\relax
\def\floatpagepagefraction{1}
\def\textpagefraction{.001}

\shorttitle{}    

\shortauthors{S.R. Arashloo}  

\title [mode = title]{Signature-Kernel One-Class Learning with Manifold Regularisation for Multivariate Time Series Anomaly Detection}  


%

\author[1]{Shervin Rahimzadeh Arashloo}[orcid=0000-0003-0189-4774]

\cormark[1]


\ead{shervin.rahimzadeh@ozyegin.edu.tr}

\ead[url]{https://sites.google.com/site/shervinrarashloo}

\credit{Conceptualization, Methodology, Formal analysis, Investigation, Software, Validation, Writing -- original draft, Writing -- review \& editing}

\affiliation[1]{organization={Department of Computer Science, 
            Faculty of Engineering,
            Özyeğin University},
            addressline={Çekmeköy}, 
            city={Istanbul},
            postcode={34794}, 
            state={Istanbul},
            country={Türkiye}}

\cortext[1]{Corresponding author}



\begin{abstract}
One-class learning is an effective methodology for time series anomaly detection, yet kernel one-class approaches often compare sequences in a pointwise fashion and ignore the geometry of normal data. In this work, we present a manifold-regularised, large-margin $\ell_p$-support vector data description ($\ell_p$-SVDD) method that analyses sequences as paths via the signature kernel. By incorporating a a graph-Laplacian penalty into the $\ell_p$-SVDD objective function and utilising the representer theorem for manifold regularisation, we illustrate that the dual task retains the functional form of the original one with a modified kernel matrix. As such, existing Frank--Wolfe solvers can be applied unchanged. We show that this new modified kernel is a valid kernel matrix and under a generic norm bound on the discriminant, the proposed model yields a smaller Rademacher-complexity bound than the baseline. Furthermore, the signature kernel which is computed through a Goursat partial differential equation and can handle irregularly sampled data and is invariant to reparametrisation has not previously been used for one-class time series anomaly detection to the best of our knowledge. Our experiments on five widely used benchmarks, evaluated without point adjustment, demonstrate that the proposed approach compares very favourably with the state of the art approaches.
\end{abstract}


\begin{highlights}
\item Manifold-regularisation in large-margin $\ell_p$-SVDD for time series anomaly detection
\item Signature kernel analyses windows as paths, handling irregular sampling
\item Proposed problem preserves the original dual form, so solvers apply unchanged
\item Rademacher analysis provides a tighter bound than the baseline
\end{highlights}


\maketitle

\section{Introduction}\label{}
The concept of regularisation has a rich mathematical history and plays a fundamental role in various machine learning algorithms. The motivation behind the idea is to encourage the model to be situated within a more confined region of all potential solutions by injecting supplementary prior knowledge or assumptions to enhance its representational capability. Among others, manifold regularisation (MR) \cite{JMLR:v7:belkin06a} has been introduced as a mechanism to leverage the geometry of the probability distribution of the data as an additional source of information for function learning. The idea is based on the assumption that if two points are close in the inherent geometry of the probability distribution that governs the production of examples, it is probable that they will share similar labels. In other words, the labels generally change gradually along the geodesics of the underlying distribution and manifold regularisation tries to benefit from such geometric smoothness assumptions to derive a better solution. 

While typically used in unsupervised or semi-supervised settings, MR can also provide advantages within a fully supervised framework. In a fully supervised setting, although labeled samples are utilised for optimisation, these labels alone may not be able to entirely capture the intricate geometric relationships present in high-dimensional or structured data, such as those present in time series or sequential observations. On the other hand, since manifold regularisation introduces a geometric prior that promotes consistency between the learned classifier and the inherent structure of the data distribution, it possesses the potential to serve as a useful mechanism to improve the performance in fully supervised settings. In the context of time series anomaly detection the idea is especially relevant as normal sequences usually adhere to smooth, low-dimensional dynamic patterns, while anomalies tend to depart from these anticipated paths. In this situation, manifold regularisation can capture the regularities in the data by applying constraints that promote conformity to the manifold of normal behaviour, thus potentially boosting anomaly detection performance.

The idea of manifold regularisation is expansive and has been applied across different learning algorithms, including deep learning approaches \cite{10144670} among others. Although deep learning-based methods have improved the performance significantly in different domains and have witnessed increasing attention in recent years, one alternative to these approaches may be considered as kernel-based algorithms \cite{10.1214/009053607000000677}. As compared with deep learning methods, kernel approaches are based on sound mathematical basis and provide theoretical guarantees on their generalisation performance. Furthermore, in the case of a scarcity of training samples, deep learning approaches offer restricted, if any, advantages. In contrast, kernel methods may be trained with much fewer training observations to achieve competitive performances in different learning scenarios. Among other kernel-based approaches, the method in \cite{RAHIMZADEHARASHLOO2024110189} presents an effective approach for anomaly detection, outperforming some other alternatives in different anomaly detection settings. Compared to other methods, the merits of the technique presented in \cite{RAHIMZADEHARASHLOO2024110189} that generalises the well-known SVDD formalism \cite{Tax2004} based on an earlier study \cite{RAHIMZADEHARASHLOO2022108930} may be summarised as follows. First, instead of a linear penalty for classification errors, the method in \cite{RAHIMZADEHARASHLOO2024110189} introduces an $\ell_{p\geq1}$-norm cost which enables the model to non-linearly penalise errors in the primal space. The norm penalty in the primal space, corresponds to a norm constraint in the dual space formulation that controls the sparsity of the solution, yielding enhanced adaptability for improved performance. Second, the method in \cite{RAHIMZADEHARASHLOO2024110189} explicitly maximises the margin between target and non-target samples, thus improving the generalisation capability of the approach. And last but not least, it solves the corresponding optimisation problem via an efficient algorithm tailored to the specific structure of the problem, ensuring improved performance.

Despite its remarkable qualities, the large-margin $\ell_p$-SVDD method in \cite{RAHIMZADEHARASHLOO2024110189} has a number of limitations. First, it does not explicitly capture and benefit from the underlying structural information of data distribution to learn an optimal classifier. This may compromise the anomaly detection performance when dealing with highly structured data with inherent correlation characteristics such as time series or sequential data. Second, conventional kernels such as the Radial Basis Function (RBF) or linear kernel used in \cite{RAHIMZADEHARASHLOO2024110189}, rely on static, pointwise comparisons and are thus not only incapable of dynamically capturing a path’s evolution, but also fail to convey a fine representation of nonlinear dependencies and higher-order interactions across multiple dimensions. Furthermore, the conventional static kernels lack invariance to time reparametrisation, a crucial property for robustness against irregular sampling and distortions along the time axis. Moreover, when the sequential data are of different lengths, these static kernel are not directly applicable, necessitating additional intermediate warping stages.

Driven by these observations, in this work, we generalise the large-margin $\ell_p$-SVDD method \cite{RAHIMZADEHARASHLOO2024110189} for time series anomaly detection. Towards this end, first, our framework exploits the geometry of the data manifold, encoding it as an additional regularisation term. This is intuitive as time series data typically incorporates densely sampled instances through time which increases the possibility of local correlation in the data, and thus, the associated labels. In this context, we elaborate on the RKHS (reproducing kernel Hilbert space) formulation of the method in \cite{RAHIMZADEHARASHLOO2024110189} and illustrate how the geometry of data manifold may be incorporated into the model through a manifold regularisation term to impose structure on the classifier learned. In particular, we illustrate how the proposed approach fits in a well-established Representer theorem presented in \cite{JMLR:v7:belkin06a} to derive the functional form of the optimal solution. By formulating the dual optimisation problem, we show that the learning problem of the proposed method resembles that of the method in \cite{RAHIMZADEHARASHLOO2024110189} with a difference in the effective kernel matrix. As such, the optimisation techniques developed in \cite{RAHIMZADEHARASHLOO2024110189} become applicable to the proposed technique.

Second, as static kernels used in \cite{RAHIMZADEHARASHLOO2024110189} fall short in capturing the complex structure of time series data, in this study, we resort to more advanced kernel functions, and specifically the signature kernel \cite{doi:10.1137/20M1366794} developed for sequential data analysis. The signature kernel is a powerful mathematical framework for time series analysis, uniquely designed to capture the rich temporal and multivariate structure of sequential data encoding both local and global dependencies within the data stream. We shall illustrate that the signature kernel is especially useful in time series anomaly detection within the proposed approach via extensive evaluations on multiple data sets. In this context, we benefit from a recent theoretical advancement representing the signature kernel as a hyperbolic PDE (partial differential equation) solution \cite{doi:10.1137/20M1366794}. Drawing on this PDE formulation, the kernel is constructed through incremental properties of the path, making the computation scalable.

Finally, using Rademacher complexities \cite{MohriRostamizadehTalwalkar18}, we conduct a theoretical analysis of the proposed method and compare it against the baseline method to illustrate the improvements achieved in terms of generalisation capability.

\subsection{Summary of contributions}
The principal contributions of this study are detailed below.
\begin{itemize}
\item Manifold-regularised large-margin $\ell_p$-SVDD for time series: we extend the method in \cite{RAHIMZADEHARASHLOO2024110189} with a graph-Laplacian penalty that promotes smoothness of the decision boundary based on the manifold of normal windows. Although manifold regularisation has been combined with SVDD before \cite{WU2023235}; to the best of our knowledge, it has not been considered for the large-margin $\ell_p$ formulation or applied to sequential anomaly detection.
\item Efficient learning: We demonstrate that the proposed regularised objective function fits the representer theorem for manifold regularisation, and its dual preserves the functional form of the original unregularised problem with a modified kernel matrix. As such, the Frank--Wolfe solver of \cite{RAHIMZADEHARASHLOO2024110189} applies without modification.
\item Generalisation analysis: Based on Rademacher complexities, we illustrate that the proposed regularised model has a smaller Rademacher complexity bound than the original baseline method.
\item Signature kernel for one-class anomaly detection: We incorporate the signature kernel evaluated via the PDE formulation of \cite{doi:10.1137/20M1366794} into the proposed approach for handling irregularly sampled multivariate data and achieving invariance to time reparametrisation. To the best of our knowledge, no earlier study has used a signature kernel for one-class detection of time series anomalies.
\item Experimental evaluation: We evaluate the proposed approach on five widely used multivariate benchmarks without point adjustment, and present the results of ablation studies of the manifold term, the kernel, the window length and the pseudo-negative samples.
\end{itemize}

\subsection{Organisation}
The remainder of the paper is arranged in the following manner. Section \ref{lit} presents a brief review of the relevant work on time series anomaly detection. Section \ref{novel} introduces the proposed manifold-regularised method along with its efficient implementation and optmisation. In Section \ref{ta}, using Rademacher complexities, we present a theoretical study of the generalisation performance of the proposed technique. Section \ref{exps} presents an experimental analysis of the proposed method along with a comparison with other methods from the literature. Finally, in Section \ref{conc}, conclusions are drawn.

\section{Related Work}
\label{lit}
While alternative classifications may exist \cite{6684530,DOMINGUES2018406}, anomaly detection models, in general, may be broadly identified as either generative \cite{11184472} or nongenerative \cite{6636290}. While in the generative group a clear connection exists between the observations and the models, nongenerative methods lack a direct association with observations. This is reflected in discriminative techniques that focus on determining the class of an input item directly. Nevertheless, the class identity information does not allow for the synthesis of a specific observation. In this sense, the main objective in discriminative models is to segment the observation space rather than modelling the underlying generative process.

Generative methods, on the other hand, try to establish a direct association between model identity and measurements. Once measurements are extracted from data, a generative model describes how those measurements are produced. Once a measurement is obtained, one can formulate a model and check if the measurement could plausibly have been generated by that model through analysing the likelihood of the observation. As an instance of generative methods, in \cite{8279425}, an auto-encoder-based method utilising long short-term memory networks is proposed to reconstruct the expected distribution of signals. For anomaly detection, a reconstruction residual score is used. Other work \cite{10.1145/3292500.3330672} uses a recurrent neural network to capture normal patterns of time series data by learning their representations and then tries to reconstruct the input and use the reconstruction probabilities for classification. In a different study \cite{9338317}, the authors try to learn the complex dependencies of multivariate time series in temporal and feature domains via a forecasting-based model and a reconstruction-based technique to derive representations through a combination of prediction and reconstruction of the data for classification. The authors in \cite{10.1145/3178876.3185996} propose an unsupervised anomaly detection method based on variational auto-encoders for time series anomaly detection. Unlike discriminative models which are directly designed for classification, the proposed generative model provides multiple outputs. For anomaly detection, the reconstruction probability of a test sample is used as the decision criterion. Other work \cite{YAO2023109084} suggests to regularise autoencoders to derive features specific to normal observations by adopting an auto-encoder-based approach. To this end, a statistical analysis on wavelet coefficients of input sequences is conducted by limiting the latent spaces to solely focus on patterns of normal sequences. The study in \cite{GIANNOULIS2023109814} proposes an encoder-decoder architecture with both implicit/explicit attention and adjustable units for predicting normality as regular patterns in sequential data based on deviations from the predictions. The work in \cite{BIANCHI2019106973} directly tries to learn compressed representations of time series data in the presence of noise and redundant information. To this end, an auto-encoder architecture utilising recurrent neural networks is proposed to generate compressed representations of variable lengths and possibly with missing data. Other method in \cite{10417809}, is designed as an encoder-decoder framework. Once temporal and spatial features are extracted, the method utilises a deep decoder to reconstruct the raw input time series. This reconstructed output is a direct synthesis of the expected normal observation. For classification, the deviation between the raw input and the reconstructed output is used. The work in \cite{10887244} presents a federated learning approach to collaboratively train learners on decentralized sequential data data. In order to deal with noisy and diverse data, the method splits each client's network into a shared branch for common patterns and a private branch for unique local traits. An attention mechanism then combines these features to reconstruct the data.

Nongenerative models do not directly evaluate the distributions of measurements. Consequently, they are unable to test the consistency of measurements against a hypothesised model. Nevertheless, nongenerative models are typically the preferred choice of practice in classification settings as they concentrate directly on classification rather than on the intermediate task of modelling the distributions of class conditional measurements. As a result of this focus on classification rather than generative process, they typically yield strong classification performance. An an example of nongenerative approaches, in \cite{NEURIPS2020_97e401a0} a temporal one-class classification approach is presented for time series anomaly detection. The method captures temporal dynamics in multiple scales through a dilated recurrent neural network. Motivated by the SVDD method \cite{Tax2004}, a one-class objective function is defined and multiple hyper-spheres obtained with a hierarchical clustering process are used for training the network for anomaly detection. The study in \cite{xu2022anomaly} proposes to use attention-based mechanisms to capture and analyse the internal associations within time series data via transformer-based architectures and tries to detect anomalies through patterns in these associations. The authors in \cite{10.14778/3514061.3514067} present an anomaly detection approach based on transformers where attention-based encoders are utilised for inference. The method facilitates feature extraction and adversarial training for improved stability. Other study \cite{Yue_Wang_Duan_Yang_Huang_Tong_Xu_2022} presents a method to learn contextual representations of time series at multiple semantic levels. To this end, a hierarchical contrasting method for capturing multi-scale contextual information and a consistency criterion for positive pair selection are used. Once effective representations are derived, a support vector classifier is used on top of the learned representations for anomaly detection. The work in \cite{10.1145/3580305.3599295} presents a multi-scale representation learning approach that deploys a dual attention structure and a contrastive loss to guide the training process to learn a representation with good discrimination potential. Unlike some other anomaly detection approaches that operate based on reconstruction residual, this approach is a self-supervised framework to learn discriminative representations to separate normal from anomalous observations. The authors in \cite{wu2023timesnet} present a nongenerative approach by focusing on learning representations of temporal variations within time series by transforming 1D sequences into 2D tensors, and then trying to make simultaneous use of inter-period and intra-period variations. Using an inception block, the method discovers multi-periodic patterns for anomaly detection. Other work in \cite{10070580} follows a multiple kernel--based approach to novelty detection for fusing multiple sources of information in the kernel space but does not consider application/extension to the sequential data. A recent study in \cite{DARBAN2025110874} proposes a nongenerative approach to time series anomaly detection using self-supervised contrastive learning. The approach presents a contrastive learning-based methodology that improves performance by injecting synthetic negative samples for training. The self-supervised scheme enables the method to derive discriminative representations for classification. Based on the merits of ensemble systems, other study \cite{NOURMOHAMMADI2026112204} tries to learn locally adaptive weights for a linear fusion of multiple classifiers but fails to consider extension to time-series data. The work in \cite{10325644} proposed a nongenerative self-supervised framework that learns representations by considering normal data as the positive class while repelling synthetic anomalies. Anomalies are then effectively classified during inference by measuring the distance between a test sample's embedding and the centroid of the normal training data. The study in \cite{10398596} presents a distribution-free method tailored for time series data. Rather than assuming an underlying data distribution, the approach predicts conditional quantiles to dynamically establish the upper and lower boundaries of normal behaviour.

The proposed approach in this study belongs to the nongenerative group and tries to directly classify samples without trying to learn the underlying generative process or probability distribution, presented next.

\section{Methodology}
\label{novel}
Suppose $\{\mathbf{x}_j\}_{j=1}^n$ are the training observations with the corresponding labels $\{y_j\}_{j=1}^n$ and $v(g(\mathbf{x}_j),y_j)$ is a loss function while $\norm{.}_{\mathcal{H}}$ denotes the norm in the Hilbert space $\mathcal{H}$. As will be discussed shortly, the proposed method uses the following theorem from \cite{JMLR:v7:belkin06a} which characterises the functional form of the optimal solution to the manifold regularisation problem in the kernel space.
\begin{thm}[The Representer theorem for manifold regularisation \cite{JMLR:v7:belkin06a}]\ \\
The solution to\\
\begin{align}
\nonumber g^{opt}(.)=&\operatorname*{\arg\,min}_{g\in \mathcal{H}}\sum_jv(g(\mathbf{x}_j),y_j)\\\nonumber&+a_1\sum_{i,j}w_{ij}(g(\mathbf{x}_i)-g(\mathbf{x}_j))^2+a_2\norm{g}_{\mathcal{H}}^2\\   =&\operatorname*{\arg\,min}_{g\in \mathcal{H}}\sum_jv(g(\mathbf{x}_j),y_j)+a_1g^\top\mathbf{L}g+a_2\norm{g}_{\mathcal{H}}^2,
\end{align}
where $w_{ij}$ is the weight of the edge between $\mathbf{x}_i$ and $\mathbf{x}_j$ in an adjacency graph and $\mathbf{L}$ denotes the graph Laplacian, admits the form $g^{opt}(.)=\sum_j\beta_j\kappa(\mathbf{x}_j,.)$ for the kernel $\kappa(.,.)$ associated with the reproducing kernel Hilbert space $\mathcal{H}$.
\label{RT}
\end{thm}

The proposed approach in this study builds on the $\ell_p$-SVDD approach presented in \cite{RAHIMZADEHARASHLOO2024110189}. In particular, we introduce a manifold regularisation term of the form $g^\top\mathbf{L}g$ into the objective function of the $\ell_p$-SVDD method to encourage smoothness of the solutions on the underlying manifold. As formally stated in following proposition, when the solution to the method in \cite{RAHIMZADEHARASHLOO2024110189} is regularised to lie on a smooth manifold, Theorem \ref{RT} may be applied to form the optimal solution.
\begin{prop}
\ \\
The objective function of the large-margin $\ell_p$-SVDD approach in the kernel space when augmented with a manifold regularisation term takes the form of Theorem \ref{RT}, and thus, its optimal solution is given as $g(.)=\sum_j\beta_j\kappa(\mathbf{x}_j,.)$.
\label{prop0}
\end{prop}

\noindent \textbf{Proof}\\
The optimisation problem associated with the large-margin $\ell_p$-SVDD approach \cite{RAHIMZADEHARASHLOO2024110189} is:
\begin{align}
\nonumber & \min_{r,\mathcal{C},\boldsymbol\zeta,\tau}r^2+c_1\sum_i\zeta_i^p+c_2\sum_l\zeta_l^p-\nu\tau^2,\\\nonumber
& \text{subject to: }\\\nonumber &\norm{\phi(\mathbf{x}_i)-\mathcal{C}}_{\mathcal{H}}^2\leq r^2-\tau^2+\zeta_i, \text{  }\zeta_i\geq0 \text{  }\forall i,\\
& \text{  } \norm{\phi(\mathbf{x}_l)-\mathcal{C}}_{\mathcal{H}}^2\geq r^2+\tau^2-\zeta_l, \text{  }\zeta_l\geq0, \text{  } \forall l,
\label{largemarg}
\end{align}
\noindent where $\mathcal{C}$ is the description centre in the Hilbert space, $r$ is the radius while $\phi(.)$ stands for a projection operator onto the Hilbert space and $\boldsymbol\zeta$ is a vector collection of the errors. In the equation above, $\tau$ controls the margin while $c_1$, $c_2$ and $\nu$ are positive trade-off parameters. In Eq. \ref{largemarg}, $i$ indexes a positive training sample while $l$ indexes a negative object. Using $-1$ and $+1$ labels for the negative and positive training samples respectively, the optimisation problem above may written as:
\begin{align}
\nonumber & \min_{r,\mathcal{C},\boldsymbol\zeta,\tau}r^2+c_1\sum_i\zeta_i^p+c_2\sum_l\zeta_l^p-\nu\tau^2,\\
& \text{subject to: } y_j\Big(\norm{\phi(\mathbf{x}_j)-\mathcal{C}}_{\mathcal{H}}^2-r^2\Big)+\tau^2\leq \zeta_j,\text{    }\zeta_j\geq0, \text{  } \forall j,
\label{comb}
\end{align}
\noindent where $j$ indexes all training samples including target and non-target objects and $y_j$ stands for a sample's label. By normalising the objects $\{\phi(\mathbf{x}_j)\}_{j=1}^n$ to have a unit magnitude in the kernel space, expanding the norm constraint in Eq. \ref{comb} yields:
\begin{align}
\nonumber & \min_{r,\mathcal{C},\boldsymbol\zeta,\tau} r^2+c_1\sum_i\zeta_i^p+c_2\sum_l\zeta_l^p-\nu\tau^2,\\
\nonumber & \text{subject to: }&\\
&y_j\Big(1-2\mathcal{C}^\top\phi(\mathbf{x}_j)+\mathcal{C}^\top \mathcal{C}-r^2\Big)+\tau^2\leq \zeta_j,\text{    }\zeta_j\geq0, \text{  } \forall j.
\end{align}
Let us suppose $\mathcal{C}^\top\mathcal{C}=\norm{\mathcal{C}}_{\mathcal{H}}^2=\lambda^2$ for an arbitrary scalar $\lambda$ and also let us define $\boldsymbol\eta\coloneqq2\mathcal{C}$. The learning problem above may then be written as
\begin{align}
\nonumber & \min_{r,\boldsymbol\eta,\boldsymbol\zeta,\tau} r^2+c_1\sum_i\zeta_i^p+c_2\sum_l\zeta_l^p-\nu\tau^2,\\
\nonumber & \text{subject to: }\\\nonumber &y_j\Big(1-\boldsymbol\eta^\top\phi(\mathbf{x}_j)+\lambda^2-r^2\Big)+\tau^2\leq\zeta_j, \text{  } \zeta_j\geq0 \text{  } \forall j,\\ &\norm{\boldsymbol\eta}_{\mathcal{H}}^2=4\lambda^2.
\end{align}
By defining $b\coloneqq 1+\lambda^2-r^2$ and $g(.)=\boldsymbol\eta^\top\phi(.)$, one obtains:
\begin{align}
\nonumber & \min_{b,\tau,g\in\mathcal{H}} -b+c^\prime\sum_j \Big(y_j\big(b-g(\mathbf{x}_j)\big)+\tau^2\Big)_+^p-\nu\tau^2,\\
& \text{subject to: } \norm{\boldsymbol\eta}_{\mathcal{H}}^2=4\lambda^2,
\end{align}
\noindent where $c^\prime=\big(c_1(1+y_j)+c_2(1-y_j)\big)/2$ and $(.)_+$ is the positive part function that returns zero for negative arguments and acts as the identity function for non-negative inputs. Following \cite{JMLR:v7:belkin06a}, for manifold regularisation, an additional term is incorporated into the objective function:
\begin{align}
\nonumber \min_{b,\tau,g\in\mathcal{H}} &-b+c^\prime\sum_j \Big(y_j\big(b-g(\mathbf{x}_j)\big)+\tau^2\Big)_+^p-\nu\tau^2\\
\nonumber&+c_3\sum_{j,k}w_{jk}(g(\mathbf{x}_j)-g(\mathbf{x}_k))^2,\\
& \text{subject to: } \norm{\boldsymbol\eta}_{\mathcal{H}}^2=4\lambda^2,
\label{someq}
\end{align}
\noindent where $w_{jk}$s denote the weights of the edges in the data adjacency graph. Since a Tikhonov and an Ivanov regularisation are equivalent \cite{JMLR:v12:kloft11a}, the problem above can be re-written as
\begin{align}
\nonumber \min_{b,\tau,g\in\mathcal{H}} &-b+c^\prime\sum_j \Big(y_j\big(b-g(\mathbf{x}_j)\big)+\tau^2\Big)_+^p-\nu\tau^2\\
 &+c_3\sum_{j,k}w_{jk}(g(\mathbf{x}_j)-g(\mathbf{x}_k))^2+c_4\norm{\boldsymbol\eta}_{\mathcal{H}}^2,
\label{E5}
\end{align}
\noindent where $c_4$ is a suitably chosen parameter. If one considers the loss function as $v(g(\mathbf{x}_j),y_j))=\min_{b,\tau}\big\{c^\prime\Big(y_j\big(b-g(\mathbf{x}_j)\big)+\tau^2\Big)_+^p-(b+\nu\tau^2)/n\big\}$ and $g = [g(x_1),\dots, g(x_n)]^\top$, the learning problem above takes the form of
\begin{align}
\min_{g\in\mathcal{H}} \sum_jv(g(\mathbf{x}_j),y_j)+c_3g^\top \mathbf{L}g+c_4\norm{g}_{\mathcal{H}}^2,
\label{MRE}
\end{align}
\noindent where $\mathbf{L}$ is the graph Laplacian. Consequently, the optimisation problem above matches that of Theorem \ref{RT}, and hence, the optimal solution to the proposed manifold-regularised anomaly detection method may be represented as $g(.)=\sum_j\beta_j\kappa(\mathbf{x}_j,.)$.$\square$
\subsection{Optimisation}
Using Proposition \ref{prop0}, the collective responses for the entire training set in the proposed manifold-regularised approach can be derived as $g=\mathbf{K}\boldsymbol\beta$ where $\mathbf{K}$ is the kernel matrix and $\boldsymbol\beta$ is a vector with elements of $\{\beta_j\}_{j=1}^n$. As a result, the optimisation problem of the proposed approach in the RKHS reads:
\begin{align}
\nonumber & \min_{r,\mathcal{C},\boldsymbol\zeta,\tau}r^2+c_1\sum_i\zeta_i^p+c_2\sum_l\zeta_l^p-\nu\tau^2+c_3 \boldsymbol\beta^\top \mathbf{K}\mathbf{L}\mathbf{K}\boldsymbol\beta,\\
&\nonumber \text{subject to: } \\
&\nonumber \norm{\phi(\mathbf{x}_i)-\mathcal{C}}_{\mathcal{H}}^2\leq r^2-\tau^2+\zeta_i, \text{  }\zeta_i\geq0, \forall i\\
&\norm{\phi(\mathbf{x}_l)-\mathcal{C}}_{\mathcal{H}}^2\geq r^2+\tau^2-\zeta_l, \text{  }\zeta_l\geq0,\text{  }, \forall l.
\label{proposedeq}
\end{align}
Next, we form the Lagrangian:
\begin{align}
\nonumber \mathcal{L} &= r^2+c_1\sum_i\zeta_i^p+c_2\sum_l\zeta_l^p-\nu\tau^2+c_3 \boldsymbol\beta^\top \mathbf{K}\mathbf{L}\mathbf{K}\boldsymbol\beta\\
\nonumber &-\sum_i \rho_i\big(r^2-\tau^2+\zeta_i-1-\norm{\mathcal{C}}_{\mathcal{H}}^2+2\mathcal{C}^\top\phi(\mathbf{x}_i)\big)\\
\nonumber &-\sum_l \rho_l\big(-r^2-\tau^2+\zeta_l+1+\norm{\mathcal{C}}_{\mathcal{H}}^2-2\mathcal{C}^\top\phi(\mathbf{x}_l)\big)\\
\nonumber &-\sum_i\mu_i\zeta_i-\sum_l\mu_l\zeta_l\\
\nonumber = & r^2+c_1\sum_i\zeta_i^p+c_2\sum_l\zeta_l^p-\nu\tau^2+c_3 \boldsymbol\beta^\top \mathbf{K}\mathbf{L}\mathbf{K}\boldsymbol\beta\\
\nonumber &-\sum_i \rho_i\big(r^2-\tau^2+\zeta_i-1-\frac{1}{4}\norm{\boldsymbol\eta}_{\mathcal{H}}^2+\boldsymbol\eta^\top\phi(\mathbf{x}_i)\big)\\
\nonumber &-\sum_l \rho_l\big(-r^2-\tau^2+\zeta_l+1+\frac{1}{4}\norm{\boldsymbol\eta}_{\mathcal{H}}^2-\boldsymbol\eta^\top\phi(\mathbf{x}_l)\big)\\
&-\sum_i\mu_i\zeta_i-\sum_l\mu_l\zeta_l
\label{lastlabel}
\end{align}
\noindent where $\rho_i$, $\rho_l$, $\mu_i$, and $\mu_l$ denote non-negative Lagrange multipliers. We assume the objects are normalised to have a unit magnitude in the kernel space and used the reparametrisation $\boldsymbol\eta=2\mathcal{C}$. According to Proposition \ref{prop0}, the optimal solution, {\em i.e.} $g(\mathbf{x})=\boldsymbol\eta^\top\phi(\mathbf{x})$, to the optimisation problem in Eq. \ref{proposedeq} can be written as $g(\mathbf{x})=\sum_j\beta_j\kappa(\mathbf{x}_j,\mathbf{x})$ using which one obtains $\boldsymbol\eta=\sum_j\beta_j\phi(\mathbf{x}_j)$, and hence, $\norm{\boldsymbol\eta}^2_{\mathcal{H}}=\boldsymbol\beta^\top\mathbf{K}\boldsymbol\beta$. Plugging $g(.)$ and $\norm{\boldsymbol\eta}^2_{\mathcal{H}}$ into the Lagrangian of Eq. \ref{lastlabel} yields:
\begin{align}
\nonumber \mathcal{L} = & r^2+c_1\sum_i\zeta_i^p+c_2\sum_l\zeta_l^p-\nu\tau^2+c_3 \boldsymbol\beta^\top \mathbf{K}\mathbf{L}\mathbf{K}\boldsymbol\beta \\
\nonumber -&\sum_i \rho_i(r^2-\tau^2+\zeta_i-1-\frac{1}{4}\boldsymbol\beta^\top\mathbf{K}\boldsymbol\beta+\boldsymbol\beta^\top\mathbf{k}_i)\\
 \nonumber -&\sum_l \rho_l(-r^2-\tau^2+\zeta_l+1+\frac{1}{4}\boldsymbol\beta^\top\mathbf{K}\boldsymbol\beta-\boldsymbol\beta^\top\mathbf{k}_l)\\
 -&\sum_i\mu_i\zeta_i-\sum_l\mu_l\zeta_l,
\end{align}
\noindent where $\mathbf{k}_i$ and $\mathbf{k}_l$ denote the $i^\mathrm{th}$ and $l^\mathrm{th}$ columns of the kernel matrix $\mathbf{K}$. Requiring the partial derivatives of the Lagrangian to vanish for minimisation w.r.t. the primal variables $r$, $\zeta_i$, $\zeta_l$, and $\tau$ yields:
\begin{subequations}
\begin{align}
     &\frac{\partial \mathcal{L}}{\partial r}=0 \hspace{5pt} \Rightarrow \sum_i\rho_i-\sum_l\rho_l=1,\label{eq:subeq0}\\
    &\frac{\partial \mathcal{L}}{\partial\zeta_i}=0 \hspace{5pt} \Rightarrow \zeta_i=(\frac{\rho_i+\mu_i}{c_1 p})^{\frac{1}{p-1}},    \label{eq:subeq1}\\
		&\frac{\partial \mathcal{L}}{\partial\zeta_l}=0 \hspace{5pt} \Rightarrow \zeta_l=(\frac{\rho_l+\mu_l}{c_2 p})^{\frac{1}{p-1}}, \label{eq:lastsub}\\
		&\frac{\partial \mathcal{L}}{\partial\tau}=0 \hspace{5pt} \Rightarrow \sum_i\rho_i+\sum_l\rho_l=\nu.\label{eq:subeq4}
\end{align}
\end{subequations}
It can be easily confirmed that Slater's condition is satisfied. As such, at the optimum, the complementary conditions hold:
\begin{subequations}
\begin{align}
    &\mu_i\zeta_i=0,\forall i\label{eq:subeq2},\\
		&\mu_l\zeta_l=0,\forall i\label{eq:subeq3},\\
    &\rho_i(r^2-\tau^2+\zeta_i-\norm{\phi(\mathbf{x}_i)-\mathcal{C}}_\mathcal{H}^2)=0, \forall i,\label{null1}\\
		&\rho_l(r^2+\tau^2-\zeta_l-\norm{\phi(\mathbf{x}_l)-\mathcal{C}}_\mathcal{H}^2)=0, \forall l\label{null2}.
\end{align}
\end{subequations}
Using Eq. \ref{eq:subeq2} and Eq. \ref{eq:subeq1} we have $\mu_i(\frac{\rho_i+\mu_i}{c_1 p})^{\frac{1}{p-1}}=0$. Since $\mu_i\geq 0$ and $\rho_i\geq 0$, we have $\mu_i=0$, and hence, using Eq. \ref{eq:subeq1} one obtains $\zeta_i=(\frac{\rho_i}{c_1 p})^{\frac{1}{p-1}}$. Using Eq. \ref{eq:subeq3} and Eq. \ref{eq:lastsub}, and following a similar analysis, we have $\mu_l=0$ and as a result $\zeta_l$ is derived as $\zeta_l=(\frac{\rho_l}{c_2 p})^{\frac{1}{p-1}}$. The Lagrangian, after re-arranging terms, would then be:
\begin{align}
\nonumber \mathcal{L} = &-c^\prime_1\norm{(\mathbf{1}+\mathbf{y})\odot\boldsymbol\rho}_q^q-c^\prime_2\norm{(\mathbf{1}-\mathbf{y})\odot\boldsymbol\rho}_q^q\\
&+c_3 \boldsymbol\beta^\top \mathbf{K}\mathbf{L}\mathbf{K}\boldsymbol\beta+\frac{1}{4}\boldsymbol\beta^\top\mathbf{K}\boldsymbol\beta-\boldsymbol\beta^\top\mathbf{K}(\mathbf{y}\odot\boldsymbol\rho),
\label{L2}
\end{align}
\noindent where $c^\prime_1=\frac{p-1}{p}(c_1p)^\frac{-1}{p-1}$, $c^\prime_2=\frac{p-1}{p}(c_2p)^\frac{-1}{p-1}$, $\mathbf{1}$ is an $n$-dimensional vector of $1$s, $q=\frac{p}{p-1}$, and $\odot$ denotes element-wise multiplication.

For optimisation of the Lagrangian w.r.t. $\boldsymbol\beta$ we require its partial derivative to vanish:
\begin{align}
\frac{\partial \mathcal{L}}{\partial \boldsymbol\beta} = 2c_3\mathbf{K}\mathbf{L}\mathbf{K}\boldsymbol\beta+\frac{1}{2}\mathbf{K}\boldsymbol\beta-\mathbf{K}(\mathbf{y}\odot\boldsymbol\rho)=\mathbf{0}.
\end{align}
Assuming that the kernel matrix is positive definite, and hence invertible, after multiplying the equation above by $\mathbf{K}^{-1}$, $\boldsymbol\beta$ is derived as
\begin{align}
\boldsymbol\beta=(2c_3\mathbf{L}\mathbf{K}+\frac{1}{2}\mathbf{I})^{-1}(\mathbf{y}\odot\boldsymbol\rho),
\end{align}
\noindent where $\mathbf{I}$ denotes an $n\times n$ identity matrix (where $n$ is the number of training samples). Denoting $(2c_3\mathbf{L}\mathbf{K}+\frac{1}{2}\mathbf{I})^{-1}=\mathbf{M}$, by plugging $\boldsymbol\beta$ into the Lagrangian of Eq. \ref{L2} one obtains:
\begin{align}
\nonumber
\mathcal{L}=&-c^\prime_1\norm{(\mathbf{1}+\mathbf{y})\odot\boldsymbol\rho}_q^q-c^\prime_2\norm{(\mathbf{1}-\mathbf{y})\odot\boldsymbol\rho}_q^q\\
\nonumber &+c_3 (\mathbf{y}\odot\boldsymbol\rho)^\top \mathbf{M}^\top\mathbf{K}\mathbf{L}\mathbf{K}\mathbf{M}(\mathbf{y}\odot\boldsymbol\rho)\\
\nonumber &+\frac{1}{4}(\mathbf{y}\odot\boldsymbol\rho)^\top\mathbf{M}^\top\mathbf{K}\mathbf{M}(\mathbf{y}\odot\boldsymbol\rho)-(\mathbf{y}\odot\boldsymbol\rho)^\top\mathbf{M}^\top\mathbf{K}(\mathbf{y}\odot\boldsymbol\rho)\\
\nonumber =&-c^\prime_1\norm{(\mathbf{1}+\mathbf{y})\odot\boldsymbol\rho}_q^q-c^\prime_2\norm{(\mathbf{1}-\mathbf{y})\odot\boldsymbol\rho}_q^q\\
&-(\boldsymbol\rho\odot\mathbf{y})^\top\mathbf{Q}(\boldsymbol\rho\odot\mathbf{y})
\label{},
\end{align}
\noindent where $\mathbf{Q}=\frac{1}{2}\mathbf{M}^\top\mathbf{K}=(4c_3\mathbf{K}\mathbf{L}+\mathbf{I})^{-1}\mathbf{K}$. The dual problem is then to maximise the Lagrangian, or equivalently, to minimise the negative Lagrangian w.r.t. $\boldsymbol\rho$ subject to the constraints given in Eq. \ref{eq:subeq0} and \ref{eq:subeq4}, that is
\begin{align}
    \nonumber &\min_{\boldsymbol\rho} \hspace{10pt}c^\prime_1\norm{(\mathbf{1}+\mathbf{y})\odot\boldsymbol\rho}_q^q+c^\prime_2\norm{(\mathbf{1}-\mathbf{y})\odot\boldsymbol\rho}_q^q\\
    \nonumber &+(\mathbf{y}\odot\boldsymbol\rho)^\top\mathbf{Q}(\mathbf{y}\odot\boldsymbol\rho),\\
    &\text{subject to: } \hspace{3pt} \mathbf{y}^\top\boldsymbol\rho=1,\hspace{3pt}\mathbf{1}^\top\boldsymbol\rho=\nu,\hspace{3pt}\boldsymbol\rho\geq0.
    \label{Dua2}
\end{align}
Once $\boldsymbol\rho$ is determined, $\boldsymbol\beta$ is computed as $\boldsymbol\beta=(2c_3\mathbf{L}\mathbf{K}+\frac{1}{2}\mathbf{I})^{-1}(\mathbf{y}\odot\boldsymbol\rho)$ to specify the optimal solution as $g(.)=\sum_j\beta_j\kappa(\mathbf{x}_j,.)$.
\\
\begin{prop}
\ \\
If $\mathbf{K}$ is positive definite and symmetric, $\mathbf{Q}$ will be also positive definite and symmetric, and thus, a valid kernel matrix.
\label{prop1}
\end{prop}
\noindent\textbf{Proof}\\
In order to prove that $\mathbf{Q}$ is positive definite and symmetric, observe that
\begin{align}
\mathbf{Q}=\Big[\mathbf{Q}^{-1}\Big]^{-1}=\Big[\mathbf{K}^{-1}(4c_3\mathbf{K}\mathbf{L}+\mathbf{I})\Big]^{-1}=\Big[4c_3\mathbf{L}+\mathbf{K}^{-1}\Big]^{-1}.
\label{prop2eq}
\end{align}
By assumption, $\mathbf{K}$ is positive definite and so is its inverse. Moreover, it is well known that the graph Laplacian is a positive semi-definite matrix. Since the sum of a positive definite and a positive semi-definite matrix is a positive definite matrix, the term inside the rightmost brackets is a positive definite matrix, and so is its inverse, {\em i.e.} $\mathbf{Q}$. Regarding symmetry, since both $\mathbf{K}^{-1}$ and $\mathbf{L}$ are symmetric matrices, the matrix inside the brackets is symmetric, and so is its inverse. Consequently, $\mathbf{Q}$ is a positive definite and symmetric matrix. $\square$

The dual of the problem in Eq. \ref{largemarg} corresponding to the large-margin method of \cite{RAHIMZADEHARASHLOO2024110189} is
\begin{align}
    \nonumber \min&_{\boldsymbol\rho} \hspace{10pt}c^\prime_1\norm{(\mathbf{1}+\mathbf{y})\odot\boldsymbol\rho}_q^q+c^\prime_2\norm{(\mathbf{1}-\mathbf{y})\odot\boldsymbol\rho}_q^q\\
    \nonumber &+(\boldsymbol\rho\odot\mathbf{y})^\top\mathbf{K}(\boldsymbol\rho\odot\mathbf{y}),\\
    &\text{subject to: } \hspace{3pt} \mathbf{y}^\top\boldsymbol\rho=1,\hspace{3pt}\mathbf{1}^\top\boldsymbol\rho=\nu, \boldsymbol\rho\geq0.
    \label{Duaother}
\end{align}

Comparing the optimisation problem associated with the proposed approach in Eq. \ref{Dua2} to its counterpart optimisation problem in Eq. \ref{Duaother} for the method of \cite{RAHIMZADEHARASHLOO2024110189}, one observes that the optimisation task for the proposed method bears similarities to that of the study in \cite{RAHIMZADEHARASHLOO2024110189} with the difference that the kernel matrix of the unregularised method ({\em i.e.} $\mathbf{K}$) is replaced by $\mathbf{Q}=(4c_3\mathbf{K}\mathbf{L}+\mathbf{I})^{-1}\mathbf{K}$ in the proposed approach. As such, the optimisation techniques developed for the approach in \cite{RAHIMZADEHARASHLOO2024110189} are directly applicable to the proposed method by considering $\mathbf{Q}$ as the kernel matrix. In this study, we use the Frank-Wolfe-based optimisation approach developed in \cite{RAHIMZADEHARASHLOO2024110189} for the optimisation of the objective function in this study by treating $\mathbf{Q}$ as the effective kernel matrix.

\subsection{Decision strategy}
In the proposed manifold-regularised approach, for classification, the distance between the hyper-sphere centre and a test object is measured and compared against the radius. The distance squared between the centre $\mathcal{C}$ and a test object $\mathbf{x}$ in the kernel space is measured as:
\begin{align}
    \norm{\phi(\mathbf{x})-\mathcal{C}}_{\mathcal{H}}^2=\kappa(\mathbf{x},\mathbf{x})-\sum_j\beta_j \kappa(\mathbf{x},\mathbf{x}_j)+\frac{1}{4}\boldsymbol\beta^\top\mathbf{K}\boldsymbol\beta.
    \label{distf}
\end{align}
For computing the radius of the hyper-spherical description, using the complementary condition in Eq. \ref{null1}, if for a positive sample $\mathbf{x}_i$ the corresponding Lagrange multiplier $\rho_i$ is not zero, we would have $r^2-\tau^2+\zeta_i-\norm{\phi(\mathbf{x}_i)-\mathcal{C}}_\mathcal{H}^2=0$. The average radius using $n_1^\prime$ such samples is $r^2=\tau^2+\frac{1}{n_1^\prime}\sum_{i|\rho_i\neq 0}\big(\norm{\phi(\mathbf{x}_i)-\mathcal{C}}_\mathcal{H}^2-\zeta_i\big)$. In a similar fashion, using Eq. \ref{null2}, if the Lagrange multiplier $\rho_l$ for a negative object $\mathbf{x}_l$ is not zero, it holds $r^2+\tau^2-\zeta_l-\norm{\phi(\mathbf{x}_l)-\mathcal{C}}_\mathcal{H}^2=0$. The average radius using $n_2^\prime$ such samples shall be $r^2=-\tau^2+\frac{1}{n_2^\prime}\sum_{l|\rho_l\neq 0}\big(\norm{\phi(\mathbf{x}_l)-\mathcal{C}}_\mathcal{H}^2+\zeta_l\big)$. Consequently, the squared radius of the hyper-sphere may be computed as
\begin{align}
\nonumber r^2 = \frac{1}{2}\Big(&\frac{1}{n_1^\prime}\sum_{i|\rho_i\neq 0}\big(\norm{\phi(\mathbf{x}_i)-\mathcal{C}}_\mathcal{H}^2-\zeta_i\big)\\
+&\frac{1}{n_2^\prime}\sum_{l|\rho_l\neq 0}\big(\norm{\phi(\mathbf{x}_l)-\mathcal{C}}_\mathcal{H}^2+\zeta_l\big)\Big).
\end{align}
Note that, in principle, the radius may be computed using a single sample $\mathbf{x}_j$ with a non-zero $\rho_j$. Nevertheless, in practice, computing the average over all such samples reduces the numerical errors. A test sample whose distance to the description centre is bigger than the radius (with respect to a certain margin) will be flagged as anomaly.

\section{Theoretical analysis}
\label{ta}
In this section, we theoretically study the generalisation error bound of the proposed manifold-regularised approach. In this context, we shall make use of the empirical Rademacher complexity \cite{MohriRostamizadehTalwalkar18} to characterise function complexity. In particular, the empirical Rademacher complexity measures, on average, how well the outputs of a class of functions may correlate with random noise. Since more complex functions are expected to have a higher capability to correlate with random noise, the Rademacher complexity provides a measure of the complexity of a family of functions.

The empirical Rademacher complexity of a family of functions $\mathcal{G}$ with respect to the sample set $\mathcal{X} = \{\mathbf{x}_j\}_{j=1}^n$ is defined as
\begin{align}
\hat{\mathcal{R}}_\mathcal{X}(\mathcal{G})=\mathbb{E}_{\boldsymbol\sigma}\Big[ \sup_{g\in\mathcal{G}}\frac{1}{n}\sum_{j=1}^n\sigma_jg(\mathbf{x}_j)\Big],
\end{align}
\noindent where $\boldsymbol\sigma = [\sigma_1, \dots, \sigma_n]^\top$ with $\sigma_j$s being independent uniform random Rademacher variables of $\{ -1,+1\}$ and $\mathbb{E}_{\boldsymbol\sigma}$ denoting the expectation with respect to $\boldsymbol\sigma$.

\begin{prop}
\ \\
For the proposed manifold regularised method, the upper bound for the empirical Rademacher complexity is strictly smaller than the baseline method in \cite{RAHIMZADEHARASHLOO2024110189}.
\label{prop2}
\end{prop}
\noindent\textbf{Proof}\\
Towards the proof, we shall first consider the theorem below which characterises kernel-based approaches' (empirical) Rademacher complexity.
\begin{thm}[Kernel-based hypotheses' Rademacher complexity \cite{MohriRostamizadehTalwalkar18}]
\ \\
Assume $\phi: \mathcal{X}\rightarrow \mathcal{H}$ a transformation that maps features from their space onto the Hilbert space with the corresponding symmetric positive-definite kernel matrix $\mathbf{K}$. Assuming $\mathcal{G}$ as a class of kernel-based functions associated with $\phi$ defined as:
\begin{align}
    \mathcal{G}=\{x\rightarrow \boldsymbol{\eta}^\top\phi(x), \norm{\boldsymbol{\eta}}_{\mathcal{H}}\leq \Lambda\},
\end{align}
for $\Lambda\geq 0$, its empirical Rademacher complexity over sample set $\mathcal{X}$ is upper-bounded as:
\begin{align}    \hat{\mathcal{R}}_\mathcal{X}(\mathcal{G})\leq\frac{\Lambda}{n}\sqrt{tr(\mathbf{K})},
\end{align}
where $tr(.)$ is the trace operator.
\label{KCRC}
\end{thm}
For the proof of Proposition \ref{prop2}, first note that:
\begin{eqnarray}
g=\mathbf{K}\boldsymbol\beta=\mathbf{K}\mathbf{M}(\mathbf{y}\odot\boldsymbol\rho)=\mathbf{Q}\big(2(\mathbf{y}\odot\boldsymbol\rho)\big)=\mathbf{Q}\boldsymbol\rho^\prime,
\end{eqnarray}
\noindent where $\boldsymbol\rho^\prime=2(\mathbf{y}\odot\boldsymbol\rho)$. That is, one may derive the responses over the training samples using $\boldsymbol\rho^\prime$ and treating $\mathbf{Q}$ as the kernel matrix. Second, note that analysing Eq. \ref{proposedeq} reveals that by setting $c_3=0$, the proposed method boils down to that of the unregularised method in \cite{RAHIMZADEHARASHLOO2024110189}. As such, we shall examine the upper bounds for the Rademacher comlexities when $c_3=0$ (representing the method in \cite{RAHIMZADEHARASHLOO2024110189}) and $c_3>0$ (corresponding to the proposed method in this work). Third, as we are interested in analysing the impact of the manifold regularisation on the Rademacher complexity independent of all the other factors, we will assume that both the proposed method and that of \cite{RAHIMZADEHARASHLOO2024110189} are similar in all aspects except for the additional manifold regularisation term. As such, for both methods we presume that the norm of Hilbert space discriminant is similarly bounded as $\norm{\boldsymbol{\eta}}_{\mathcal{H}}\leq \Lambda$. As a result, the difference between the Rademacher complexities of the two methods can be solely characterised using the traces of the corresponding kernel matrices:
\begin{align}
\nonumber
&\hat{\mathcal{R}}_\mathcal{X}(\mathcal{G})\leq\frac{\Lambda}{n}\sqrt{tr(\mathbf{K})},\\
&\hat{\mathcal{R}}_\mathcal{X}(\mathcal{G}_{MR})\leq\frac{\Lambda}{n}\sqrt{tr(\mathbf{Q})},
\end{align}
\noindent where $\mathbf{K}$ and $\hat{\mathcal{R}}_\mathcal{X}(\mathcal{G})$ respectively denote the kernel matrix and the empirical Rademacher complexity associated with the $\ell_p$-SVDD approach of \cite{RAHIMZADEHARASHLOO2024110189} while $\mathbf{Q}$ and $\hat{\mathcal{R}}_\mathcal{X}(\mathcal{G}_{MR})$ correspond to those of the proposed method. As illustrated in Eq. \eqref{prop2eq}, $\mathbf{Q}$ can be expressed as
\begin{align}
\mathbf{Q}=\Big[4c_3\mathbf{L}+\mathbf{K}^{-1}\Big]^{-1}.
\end{align}
Considering the matrix inside the brackets above and observing the fact that the trace of a sum is the sum of traces, we have:
\begin{align}
tr(4c_3\mathbf{L}+\mathbf{K}^{-1})=4c_3tr(\mathbf{L})+tr(\mathbf{K}^{-1})>tr(\mathbf{K}^{-1}),
\label{trineq}
\end{align}
\noindent where the last inequality is true when $c_3>0$ since the trace of the Laplacian matrix for an undirected graph with postive edge weights and without self-loops (the case in this study) is strictly positive for a graph with at least one edge. Let us assume $\delta_k(.)$ returns the $k^\mathrm{th}$ smallest eigenvalue of a matrix. Based on the Weyl's inequality \cite{Weyl1912}, we have 
\begin{align}
\nonumber \delta_k(4c_3\mathbf{L}+\mathbf{K}^{-1})\geq \delta_k(\mathbf{K}^{-1})+\delta_1(4c_3\mathbf{L})=1/\delta_k(\mathbf{K})\\
+4c_3\delta_1(\mathbf{L})\geq1/\delta_k(\mathbf{K}),
\label{eqt}
\end{align}
where $\delta_1(\mathbf{L})$ represents the lowest eigenvalue of the graph Laplacian matrix $\mathbf{L}$ and the last inequality is due to positive semi-definiteness of $\mathbf{L}$. We argue that:
\begin{align}
\exists k: \delta_k(4c_3\mathbf{L}+\mathbf{K}^{-1})>1/\delta_k(\mathbf{K}).
\label{ex}
\end{align}
To prove the statement, we shall use contradiction. Let us assume $\delta_k(4c_3\mathbf{L}+\mathbf{K}^{-1})=1/\delta_k(\mathbf{K}), \forall k$. In this case, one would have $\sum_k \delta_k(4c_3\mathbf{L}+\mathbf{K}^{-1})=\sum_k1/\delta_k(\mathbf{K})$. The trace of a matrix is equal to the sum of its eigenvalues which yields $tr(4c_3\mathbf{L}+\mathbf{K}^{-1})=tr(\mathbf{K}^{-1})$. This result, however, contradicts the fact that $tr(4c_3\mathbf{L}+\mathbf{K}^{-1})>tr(\mathbf{K}^{-1})$ as illustrated in Eq. \ref{trineq}. As such, the assumption $\delta_k(4c_3\mathbf{L}+\mathbf{K}^{-1})=1/\delta_k(\mathbf{K}), \forall k$ is incorrect, and hence, Eq. \ref{ex} holds. Next, using Eq. \ref{eqt}, we have
\begin{align}
\sum_k 1/\delta_k(4c_3\mathbf{L}+\mathbf{K}^{-1})\leq \sum_k\delta_k(\mathbf{K}),
\end{align}
and as a result, we have $tr([4c_3\mathbf{L}+\mathbf{K}^{-1}]^{-1})\leq tr(\mathbf{K})$. Nevertheless, based on Eq. \ref{ex}, there exists at least one eigenvalue $\delta_k(4c_3\mathbf{L}+\mathbf{K}^{-1})$ such that $1/\delta_k(4c_3\mathbf{L}+\mathbf{K}^{-1})<\delta_k(\mathbf{K})$. Consequently, we have $tr([4c_3\mathbf{L}+\mathbf{K}^{-1}]^{-1})=tr(\mathbf{Q})<tr(\mathbf{K})$ which implies that for the proposed method ($c_3>0$) the upper bound for the empirical Rademacher complexity is strictly smaller than that of \cite{RAHIMZADEHARASHLOO2024110189}. $\square$

A lower bound for the Rademacher complexity in the proposed manifold-regularised approach is intuitively expected as the proposed method imposes additional regularisation on the solution. Thus, the proposed approach is expected to yield a smoother function on the underlying manifold with a reduced complexity. A function with a lower Rademacher complexity is more likely to yield a lower classification error as formally stated in the next proposition.

\begin{prop}
\ \\
For identical margin and training error rates, the proposed manilfold-regularised approach has a reduced upper bound for misclassification probability of a test sample compared to the baseline method of \cite{RAHIMZADEHARASHLOO2024110189}.
\label{prop3}
\end{prop}

\noindent\textbf{Proof}\\
According to the analysis conducted in \cite{RAHIMZADEHARASHLOO2024110189}, with confidence greater than $1-\gamma$, the probability of incorrectly classifying a test point for the method of \cite{RAHIMZADEHARASHLOO2024110189} is bounded as
\begin{align}
    P[y\big(g(\mathbf{x})-r^2\big)>0]\leq \frac{1}{n\tau^{2p}}\norm{\boldsymbol\zeta}_p^p+\hat{\mathcal{R}}_\mathcal{X}(\mathcal{G})+3\sqrt{\frac{\ln(2/\gamma)}{2n}},
\end{align}
\noindent where $\mathbf{x}$ denotes a test data with the ground truth label $y$ and $\hat{\mathcal{R}}_\mathcal{X}(\mathcal{G})$ stands for the empirical Rademacher complexity. Following a similar analysis as that of \cite{RAHIMZADEHARASHLOO2024110189} and omitting the intermediate steps, the probability of misclassification for the proposed manifold-regularised approach shall be
\begin{align}
    \nonumber &P[y\big(g(\mathbf{x})-r^2\big)>0]\leq\\
    &\frac{1}{n\tau^{2p}}\norm{\boldsymbol\zeta}_p^p+\hat{\mathcal{R}}_\mathcal{X}(\mathcal{G}_{MR})+3\sqrt{\frac{\ln(2/\gamma)}{2n}},
\end{align}
\noindent where $\hat{\mathcal{R}}_\mathcal{X}(\mathcal{G}_{MR})$ represents the (empirical) Rademacher complexity of the proposed manifold-regularised method.
According to Proposition \ref{prop2}, the upper bound for $\hat{\mathcal{R}}_\mathcal{X}(\mathcal{G}_{MR})$ is lower than the upper bound for $\hat{\mathcal{R}}_\mathcal{X}(\mathcal{G})$. As such, for identical margin and training error rates, the upper bound for the probability of classification error in the proposed method is lower than that of \cite{RAHIMZADEHARASHLOO2024110189}. $\square$

While the proposition above theoretically sets out the superiority of the proposed method with regards to the probability of classification error compared to the method of \cite{RAHIMZADEHARASHLOO2024110189}, in the next section, we experimentally examine the advantages offered by the proposed manifold-regularised method on multiple datasets for time series anomaly detection.

\section{Experiments}
\label{exps}
This section presents and discusses the outcomes of an experimental examination of the proposed approach across several datasets, along with a comparison to the state-of-the-art approaches. The remainder of this section is organised as detailed next.
\begin{itemize}
\item Section \ref{dets} presents specifics of our implementation;
\item In Section \ref{dats}, we briefly introduce the datasets employed in this study;
\item Section \ref{ress} presents the results of an experimental assessment of the proposed technique on multiple widely used multidimensional time series anomaly detection datasets and provides a comparison against state-of-the-art methods;
\item And finally, Section \ref{abs} presents an ablation study, analysing the impacts of each component in the proposed approach.
\end{itemize}

\subsection{Implementation details}
\label{dets}
In time series anomaly detection, the time series is divided into overlapping windows with a stride of one time step, and the goal is to detect anomalies in thus obtained windows. Following the majority of existing work on time series anomaly detection, we use a length of 100 for the windows. Nevertheless, we shall also analyse the effect of changing the window size on the performance of the proposed approach in Section \ref{abs}.

In the proposed method, before constructing the kernel, we normalise all time series by rescaling the data in a way that the maximum value across all dimensions and times in each dataset is 1. In this study, we utilise the signature kernel, a positive-definite kernel specifically designed for the analysis of complex sequential data streams \cite{doi:10.1137/20M1366794}. The signature kernel can handle irregularly sampled, multivariate time series, transforming raw data into a feature set. Notably, traditional methods highlighted in the literature, like the dynamic time warping/global alignment kernel \cite{10.5555/3104482.3104599}, typically fail to produce positive definite kernel matrices when dealing with time series of varying lengths. Conversely, the truncated signature kernel \cite{JMLR:v20:16-314} only approximates the true signature kernel, which requires significant computational resources for a sufficiently accurate approximation. In this regard, the approach outlined in \cite{doi:10.1137/20M1366794} formulates the signature kernel as the solution to a partial differential equation (PDE), enabling efficient computation, which will be briefly discussed next. 

Consider a continuously differentiable time series $\mathbf{x}$ over the domain $[u,u^\prime]$. The path $\mathbf{x}$'s signature restricted to the sub-interval $[u,l]$ for $l\in[u,u^\prime]$, denoted as $S(\mathbf{x})_{l}$, is defined in terms of an integral equation:
\begin{eqnarray}
S(\mathbf{x})_{l} = \mathbb{I}+\int_{s=u}^lS(\mathbf{x})_{s}\otimes dx_s,
\end{eqnarray}
where $S(\mathbf{x})_u = \mathbb{I}=(1,0,0,\dots)$ and $\otimes$ indicates the tensor product. The signature kernel represents a reproducing kernel that measures the similarity between a pair of paths $\mathbf{x}$ and $\mathbf{y}$ in terms of their signatures:
\begin{eqnarray}
    \kappa_{l,m}(\mathbf{x},\mathbf{y})= S(\mathbf{x})_l^\top S(\mathbf{y})_m,
\end{eqnarray}
\noindent where $S(\mathbf{y})_m$ denotes the signature of path $\mathbf{y}$ (defined over the interval $[v,v^\prime]$) restricted to the sub-interval $[v,m]$ for any $m\in[v,v^\prime]$. In \cite{doi:10.1137/20M1366794}, it is shown that if $\mathbf{X}$ and $\mathbf{Y}$ are continuous kernel space representations of continuously differentiable paths $\mathbf{x}$ and $\mathbf{y}$ with variations which are bounded, the following PDE for the computation of the signature kernel for $\mathbf{X}$ and $\mathbf{Y}$ holds:
\begin{eqnarray} \label{eq1}
    \nonumber &\frac{\partial^2 \kappa_{l,m}({\mathbf{X},\mathbf{Y}})}{\partial l \partial m} = \big({\mathbf{X}^._l}^\top \mathbf{Y}^._{m}\big) \kappa_{l,m}({\mathbf{X},\mathbf{Y}}),\\ &\kappa_{u, .}({\mathbf{X},\mathbf{Y}}) = \kappa_{., v}(\mathbf{X},\mathbf{Y}) = 1,
\end{eqnarray}
\noindent where $\mathbf{X}^._l$ and $\mathbf{Y}^._m$ denote the derivatives of $\mathbf{X}$ and $\mathbf{Y}$ at time $l$ and $m$, respectively, which for piecewise linear paths can be approximated using first-order finite differences. Furthermore, using a forward finite difference scheme to approximate the differential operator and assuming that the domains of $\mathbf{X}$ and $\mathbf{Y}$ are partitioned as $\{u = u_0<u_1<\dots<u_{n_1-1}<u_{n_1}=u^\prime\}$ and $\{v=v_0<v_1<\dots<v_{n_2-1}<v_{n_2}=v^\prime\}$ where $n_1$ denotes the length of $\mathbf{X}$ and $n_2$ represents that of $\mathbf{Y}$, the following recursive formula for $i=0,\dots,n_1-1$ and $j=0,\dots,n_2-1$ for computing the signature kernel between $\mathbf{X}$ and $\mathbf{Y}$ with the initial conditions of $\kappa_{u_0, .}(\mathbf{X},\mathbf{Y}) = \kappa_{.,v_0}(\mathbf{X},\mathbf{Y}) = 1$ may be applied:
\begin{flalign*}
\nonumber &{\kappa_{u_{i+1},v_{j+1}}}(\mathbf{X},\mathbf{Y})=\\
\nonumber &{\kappa_{u_{i+1},v_j}}(\mathbf{X},\mathbf{Y}) + {\kappa_{u_i,v_{j+1}}}(\mathbf{X},\mathbf{Y}) +(C-1){\kappa_{u_i,v_j}}(\mathbf{X},\mathbf{Y}),\\
\end{flalign*}
where $C=\theta_{u_{i+1},v_{j+1}}(\mathbf{x}, \mathbf{y}) - \theta_{u_{i},v_{j+1}}(\mathbf{x}, \mathbf{y}) - \theta_{u_{i+1},v_{j}}(\mathbf{x}, \mathbf{y}) + \theta_{u_{i},v_{j}}(\mathbf{x}, \mathbf{y})$ and $\theta(.,.)$ stands for the static kernel satisfying $\theta(\mathbf{x},\mathbf{y})=\mathbf{X}^\top\mathbf{Y}$. The signature kernel need not yield unit-length objects in the feature space. To obtain unit-length samples in the kernel space, in this work, we normalise the kernel function as
\begin{eqnarray}
\label{norma}
 \kappa_{l,m}(\mathbf{X},\mathbf{Y})={\kappa_{l,m}(\mathbf{X},\mathbf{Y})}/{\sqrt{\kappa_{l,l}(\mathbf{X},\mathbf{X}) \cdot \kappa_{m,m}(\mathbf{Y},\mathbf{Y})}}.
\end{eqnarray}

For data augmentation, in this work, we use pseudo-anomaly generation. The technique used to generate pseudo-negative samples is that employed in \cite{DARBAN2025110874}. Although the technique may not be able to generate every possible type of anomaly, nevertheless, in practice it has been observed to be able to cover common time series anomaly types. Note that in the proposed approach only pseudo-negative samples and no real negative samples are used for training.

The $\nu$ parameter in the proposed method is selected from $\{1.1, 2, 4, 10\}$ while $q$ is selected from $\{16/15, 8/7, 4/3, 2, 4, 8, 16\}$ and $c_3$ from $\{1/4, 10/4, 100/4\}$. On each dataset, we randomly divide the given training set into two non-overlapping sets to be used as the positive train and validation sets. The negative train and validation sets are then generated using the pseudo-negative sample generation method used in \cite{DARBAN2025110874}. All the parameters of the proposed method are set on the validation subset of each dataset. The static kernel used in the signature kernel is that of an RBF kernel whose width is tuned to the average pairwise Euclidean distance between positive training samples. In order to learn graph adjacency edge weights, the method presented in \cite{9591377} is used.

To facilitate a fair comparison with the existing approaches, the performance metrics used to evaluate the proposed method are precision, recall, F1 score, and the area under the precision–recall curve (AU-PR). Following \cite{DARBAN2025110874}, in this study, we do not use Point Adjustment for performance reporting. Although popular among some studies, in \cite{DARBAN2025110874} it is found that PA leads to an over-estimation of anomaly detection approaches for time series and biases the evaluation results. In order to calculate the F1 score for benchmark databases containing multiple sub-datasets, as suggested in \cite{DARBAN2025110874}, we use the number of true negatives (TN), true positives (TP), false negatives (TN) and false positives (FP) on each sub-database and sum them up to obtain an overall confusion matrix for the entire database. The cumulative confusion matrix is then utilised to derive precision, recall, and F1 score. In addition to the metrics above, we make use of the G-mean metric as defined in \cite{Kubt1997AddressingTC} and suggested in \cite{FATEMIFAR2022108500} for anomaly detection to better gauge the performances in the ablation study. The G-mean is defined as
\begin{align}
\mathrm{G-Mean} = \sqrt{\mathrm{Sensitivity}\times\mathrm{Specificity}},
\end{align}
\noindent where the Sensitivity and Specificity measures are defined as $\mathrm{Sensitivity}=TP/(TP + FN)$ while $\mathrm{Specificity}=TN/(TN+FP)$. Consequently, the accuracy of both the target and the abnormal classes is considered concurrently using the G-Mean metric above. One attractive feature of G-Mean is its ability in providing a realistic assessment of performance, particularly when dealing with highly imbalanced datasets.

\subsection{Datasets}
\label{dats}
The datasets used in this study represent the most widely used time series anomaly detection databases, briefly introduced next.
\begin{itemize}
\item NASA Datasets - MSL \cite{9537291,10.1145/3219819.3219845}: Mars Science Laboratory is an expert-labeled telemetry anomaly dataset from the NASA Curiosity rover. It incorporates anomalous data of incident reports corresponding to a monitoring system on the spacecraft;

\item NASA Datasets - SMAP \cite{10.1145/3219819.3219845}: Soil Moisture Active-Passive is a collection of telemetry data from NASA's SMAP satellite. The dataset contains multivariate time series data from the satellite's operational systems.

\item Server Machine Dataset - SMD \cite{10.1145/3292500.3330672} is made up from 28 different machines represented as 28 subsets to be evaluated separately, with normal data obtained from an Internet company.

\item Secure water treatment - SWaT \cite{7469060} represents data from a water treatment platform obtained from 51 sensors and actuators over 11 days of continuous operation: seven days under normal operation and four days with attack settings, incorporating 41 anomalous samples representing a wide range of attacks created over the last four days.

\item Water distribution testbed - WADI \cite{10.1145/3055366.3055375} is a small-scale, high-fidelity, industry-compliant emulation of a modern water distribution facility equipped with capabilities to simulate physical attacks such as water leakage, malicious chemical injections, and water hammers. It incorporates a total of 123 sensors and actuators and extends over a period of sixteen days, where anomalous objects are in the last 2 days.
\end{itemize}

\begin{table}[t]
\centering
\footnotesize
\caption{Comparison of different time series anomaly detection methods in terms of Precision (Prec.), Recall (Rec.), F1, and AU-PR (\%) on multivariate time series databases. The mean rank is based on the F1 metric using Skillings–Mack Test \cite{Skillings} ($p=1.89 \times 10^{-4}$). ``-" indicates an unavailable result.\\}
\label{tab:results}
\begin{adjustbox}{max width=1\columnwidth} \begin{tabular}{@{}llcccccc@{}}
\toprule
\textbf{Method} & & \textbf{MSL} & \textbf{SMAP} & \textbf{SMD} & \textbf{SWaT} & \textbf{WADI} & \textbf{Mean rank} \\
\midrule
\multirow{4}{*}{\begin{tabular}{@{}l@{}}OmniAnom \cite{10.1145/3292500.3330672} \\ \end{tabular}}
  & AU-PR & 14.9 & 11.5 & 36.5 & 71.3 & 12.0 & \\
	& F1 & 24.3 & 32.5 & 45.9 & 76.2 & 22.8 & 7.2\\  
	& Prec. & 14.0 & 19.6 & 30.6 & 90.6 & 13.1 & \\
  & Rec. & 90.8 & 94.2 & 91.2 & 65.8 & 86.7 & \\  
\addlinespace
\hline
\multirow{4}{*}{\begin{tabular}{@{}l@{}}LSTM-VAE \cite{8279425} \\ \end{tabular}} 
  & AU-PR & 28.5 & 25.8 & 39.5 & 68.5 & 3.9 & \\
	& F1 & 40.7 & 43.7 & 29.8 & 72.3 & 11.2 & 7.9\\  
	& Prec. & 27.2 & 29.6 & 20.4 & 97.0 & 5.9 & \\
  & Rec. & 80.8 & 83.0 & 54.9 & 57.6 & 100.0 & \\  
\addlinespace
\hline
\multirow{4}{*}{\begin{tabular}{@{}l@{}}THOC \cite{NEURIPS2020_97e401a0} \\ \end{tabular}}
  & AU-PR & 23.9 & 19.5 & 10.7 & 53.7 & 10.3 & \\
	& F1 & 30.9 & 32.7 & 16.7 & 63.8 & 15.7 & 9.0\\  
	& Prec. & 19.3 & 20.3 & 9.9 & 54.5 & 10.1 & \\
  & Rec. & 77.1 & 82.9 & 53.0 & 76.8 & 35.0 & \\  
\addlinespace
\hline
\multirow{4}{*}{\begin{tabular}{@{}l@{}}MTAD-GAT \cite{9338317} \\ \end{tabular}}
  & AU-PR & 33.5 & 33.9 & 40.1 & 9.5 & 8.4 & \\
	& F1 & 47.3 & 51.8 & 34.7 & 24.2 & 12.5 & 7.0\\
	& Prec. & 35.5 & 37.8 & 24.7 & 13.8 & 7.0 & \\
  & Rec. & 77.6 & 82.3 & 58.5 & 95.8 & 58.3 & \\  
\addlinespace
\hline
\multirow{4}{*}{\begin{tabular}{@{}l@{}}AnomTran \cite{xu2022anomaly} \\ \end{tabular}}
  & AU-PR & 23.6 & 26.4 & 27.3 & 68.1 & 4.0 & \\
	& F1 & 34.4 & 40.7 & 30.4 & 73.7 & 11.3 & 7.8\\
	& Prec. & 21.8 & 26.6 & 20.6 & 97.1 & 6.0 & \\
  & Rec. & 82.3 & 86.0 & 58.2 & 59.4 & 96.0 & \\  
\addlinespace
\hline
\multirow{4}{*}{\begin{tabular}{@{}l@{}}TS2Vec \cite{Yue_Wang_Duan_Yang_Huang_Tong_Xu_2022} \\ \end{tabular}}
  & AU-PR & 13.2 & 14.8 & 11.3 & 13.6 & 5.7 & \\
	& F1 & 29.9 & 37.1 & 17.2 & 26.1 & 11.9 & 9.8\\
	& Prec. & 18.3 & 23.5 & 10.3 & 15.3 & 6.5 & \\
  & Rec. & 81.7 & 88.2 & 52.9 & 87.4 & 71.2 & \\  
\addlinespace
\hline
\multirow{4}{*}{\begin{tabular}{@{}l@{}}TranAD \cite{10.14778/3514061.3514067} \\ \end{tabular}}
  & AU-PR & 27.8 & 28.7 & 41.2 & 19.2 & 3.9 & \\
	& F1 & 42.8 & 47.1 & 36.0 & 31.9 & 11.2 & 7.5\\
	& Prec. & 29.5 & 33.6 & 26.4 & 19.2 & 5.9 & \\
  & Rec. & 77.6 & 78.8 & 56.6 & 79.6 & 100.0 & \\  
\addlinespace
\hline
\multirow{4}{*}{\begin{tabular}{@{}l@{}}TimesNet \cite{wu2023timesnet} \\ \end{tabular}}
  & AU-PR & 28.3 & 20.8 & 38.5 & 8.3 & 8.4 & \\
	& F1 & 35.7 & 40.1 & 38.8 & 21.6 & 14.4 & 8.1\\
	& Prec. & 22.5 & 25.8 & 24.5 & 12.1 & 13.3 & \\
  & Rec. & 86.2 & 89.9 & 54.7 & 100.0 & 15.6 & \\  
\addlinespace
\hline
\multirow{4}{*}{\begin{tabular}{@{}l@{}}DCdetector \cite{10.1145/3580305.3599295} \\ \end{tabular}}
  & AU-PR & 12.9 & 12.4 & 4.3 & 12.6 & 12.1 & \\
	& F1 & 22.7 & 27.5 & 8.2 & 21.6 & 24.7 & 10.7\\
	& Prec. & 12.8 & 16.0 & 4.3 & 12.1 & 14.1 & \\
  & Rec. & 95.7 & 96.1 & 99.6 & 99.9 & 96.8 & \\  
\addlinespace
\hline
\multirow{4}{*}{\begin{tabular}{@{}l@{}}CARLA \cite{DARBAN2025110874}\end{tabular}}
  & AU-PR & 50.1 & 44.8 & 50.7 & 68.1 & 12.6 & \\
	& F1 & 52.2 & 52.9 & 51.1 & 72.0 & 29.5 &3.8\\
	& Prec. & 38.9 & 39.4 & 42.7 & 98.8 & 18.5 & \\
  & Rec. & 79.5 & 80.4 & 63.6 & 56.7 & 73.1 & \\  
	\addlinespace
    \hline
\multirow{4}{*}{\begin{tabular}{@{}l@{}}CTAD \cite{10325644}\end{tabular}}
  & AU-PR & 42.3 & 32.3 & - & 80.2 & 39.9 & \\
	& F1 & 57.6 & 52.8 & - & 84.1 & 48.1 &2.75\\
	& Prec. & 57.5 & 49.2 & - & 97.1 & 79.2 & \\
  & Rec. & 57.8 & 57.0 & - & 74.2 & 34.6 & \\  
	\addlinespace
\hline
\multirow{4}{*}{\begin{tabular}{@{}l@{}} HCR-ADAAD \cite{10417809}\end{tabular}}
  & AU-PR & - & - & - & - & - & \\
	& F1 & 91.8 & 91.7 & 91.9 & - & - &2\\
	& Prec. & 89.7 & 90.4 & 90.8 & - & - & \\
  & Rec. & 93.7 & 93.1 & 92.8 & - & - & \\  
	\addlinespace
\hline
\multirow{4}{*}{\begin{tabular}{@{}l@{}}This work \end{tabular}}
  & AU-PR & 92.1 & 87.1  & 97.5 & 89.4 & 94.9 & \\
	& F1 & 95.7 & 95.8 & 98.1 & 93.5 & 97.0 &1\\
	& Prec. &  92.3 & 92.4 &  96.5 & 87.9  & 94.2 & \\
  & Rec. & 99.5 & 99.3 & 99.8 &  100.0 & 100.0 & \\  
\bottomrule
\end{tabular}\end{adjustbox}
\end{table}

\subsection{Results}
\label{ress}
The experimental results of an evaluation of the proposed approach on multiple widely used multivariate time series datasets along with a comparison with the state-of-the-art methods from the literature are presented in Table \ref{tab:results}. For a fair comparison, since in the proposed approach point adjustment is not used for performance reporting, we compare the proposed method against those from the literature following the same strategy. Note that in addition to the methods in Table \ref{tab:results}, there are other methods in literature such as those in \cite{FU2024119978,XU2026122789,ZHANG2026122680,CHEN2024120222}. Nevertheless, as in this study we do not use point adjustment for detection, the proposed approach is not directly comparable with the methods that use point adjustment for performance reporting.

From the table the following observations are in order. On all datasets, the proposed method achieves the best performance both in terms of F1 as well as AU-PR measures. The improvements in the performance achieved by the proposed approach compared to the second best performing method ({\em i.e.} HCR-ADAAD \cite{10417809}) is large, reaching up to $~7\%$ in F1 on some datasets such as MSL while being $\approx 40\%$ better compared to some other recent methods such as CARLA \cite{DARBAN2025110874}. The relatively higher F1 and AU-PR measures of the proposed approach clearly illustrates its superior overall performance as compared with other methods. The proposed method also offers both a high precision as well as a high recall rate. In the rightmost column of Table \ref{tab:results}, so as to provide an overall statistical comparison of different methods on all the databases, we provide the results of an average ranking of different methods based on the F1 rates using the Skillings–Mack test. The Skillings–Mack test is a generalisation of the Friedman's test when performance data for some methods on some datasets are missing. The results of this statistical analysis confirms that the proposed approach outperforms all approaches, ranking first among other competitors. It is worth reiterating that the in the proposed approach real negative training samples are not used. The only negative training objects deployed are pseudo-negative samples generated by the approach presented in \cite{DARBAN2025110874}. It is expected that a more general pseudo-negative generation mechanism may boost the performance even further. 

\subsection{Ablation study}
\label{abs}
In this section, we present an experimental analysis of the impacts of manifold regularisation, varying time window length, and the utility of using pseudo-negative samples on the performance.

\subsubsection{Effect of manifold regularisation}
For analysing the merits of manifold regularisation on time series anomaly detection, the performance of the proposed approach is compared against the large-margin $\ell_p$-SVDD method of \cite{RAHIMZADEHARASHLOO2024110189} which does not utilise manifold regularisation, {\em i.e.} against the case where $c_3=0$ (see Eq. \ref{proposedeq}). We compare the two methods using the AU-PR and G-mean metrics. While the AU-PR provides an overall estimate of the performance independent of any threshold, the G-mean metric offers a balanced estimate of the accuracies of both the negative and the positive classes. The results of this comparison is presented in Table \ref{impact}. The following observations may be made from the table. On all databases, the manifold regularisation improves the performance as compared with the baseline (un-regularised) approach. This is expected as by incorporating additional information regarding the data manifold, the learning algorithm is expected to adapt to the inherent structure of the data. While on some databases the improvements acheived may be moderate, on some datasets the improvemnts are huge, reaching reaching approximately $5\%$ in terms of G-mean. One expects the improvements obtained via manifold regularisation be affected by two factors. First, the inherent structure relevant to the data; {\em i.e.} whether the data has a strong structure or not to be deployed for regularisation. The second factor is that the effectiveness of the utilised approach to capture correlations and enforce smoothness. While the first factor may not be controlled by the examiner, the utility of different mechanisms for capturing the manifold structure serves as an ongoing research direction.

\begin{table}[t!]
\renewcommand{\arraystretch}{1}
\caption{The effect of manifold regularisation for time series anomaly detection on multiple data sets in terms of AU-PR (denoted as ``A'') and G-mean (denoted as ``G'') (\%).}
\label{impact}
\centering
\begin{adjustbox}{max width=1\columnwidth}
\begin{tabular}{lccccccccccccccc}
\hline
Dataset&\multicolumn{2}{c}{MSL}&\multicolumn{2}{c}{SMAP}&\multicolumn{2}{c}{SMD}&\multicolumn{2}{c}{SWAT}&\multicolumn{2}{c}{WADI}\\
\cmidrule(lr){2-3} \cmidrule(lr){4-5} \cmidrule(lr){6-7}\cmidrule(lr){8-9}\cmidrule(lr){10-11}\cmidrule(rr){12-13}\\
&A&G&A&G&A&G&A&G&A&G&\\
\hline
Manifold regularisation &92.1&66.6&87.1&70.4&97.5&68.5&89.4&52.8&94.9&54.5\\
No manifold regularisation &90.2&64.2&84.7&67.4&96.9&66.3&89.1&51.2&93.5&49.1\\
\hline
\end{tabular}\end{adjustbox}
\end{table}

\subsubsection{Effect of kernel function}
In this section, we study the merits of the signature kernel by comparing it against two commonly used static kernels, namely, the Gaussian (RBF) and the linear kernels. For the Gaussian and linear kernels, each window is flattened into a vector, and the kernel function is computed between the resulting vectors. The width of the Gaussian kernel is set to the average pairwise Euclidean distance between positive training data. For all kernels we apply normalization the kernel space based on Eq. \ref{norma}. The results are reported in Table \ref{kern} in terms of the AU-PR and G-mean metrics. As may be observed from the table, the signature kernel yields a better performance as compared with the static kernels commonly used in kernel methods. More specifically, as static kernels rely on pointwise comparisons, they cannot model the temporal ordering of the observations, the evolution of the sequential path. In contrast, the signature kernel measures the similarities between paths based on their signatures, which captures both local and global dependencies. As such, the signature kernel provides a better representation of normal dynamics, and hence a better separation between normal and anomalous windows. Amon the two static kernels, the Gaussian kernel typically performs better, thanks to its nonlinear modelling capacity between samples.

\begin{table}[t!]
\renewcommand{\arraystretch}{1}
\caption{The effect of the kernel function on the performance in terms of AU-PR (denoted as ``A'') and G-mean (denoted as ``G'') (\%).}
\label{kern}
\centering
\begin{adjustbox}{max width=1\columnwidth}
\begin{tabular}{lcccccccccc}
\hline
Dataset&\multicolumn{2}{c}{MSL}&\multicolumn{2}{c}{SMAP}&\multicolumn{2}{c}{SMD}&\multicolumn{2}{c}{SWaT}&\multicolumn{2}{c}{WADI}\\
\cmidrule(lr){2-3} \cmidrule(lr){4-5} \cmidrule(lr){6-7}\cmidrule(lr){8-9}\cmidrule(lr){10-11}\\
&A&G&A&G&A&G&A&G&A&G\\
\hline
Linear kernel &88.9&59.4&84.8&69.3&92.9&61.5&82.6&47.7&89.2&50.4\\
Gaussian kernel &90.1&60.3&84.7&68.1&93.8&64.7&84.8&49.1&90.5&51.2\\
Signature kernel &92.1&66.6&87.1&70.4&97.5&68.5&89.4&52.8&94.9&54.5\\
\hline
\end{tabular}\end{adjustbox}
\end{table}

\subsubsection{Effect of window length}
In this section, we investigate the impact of the window length on the anomaly detection performance. For this purpose, we use window sizes of $20$, $50$, $100$, and $200$ and compare the performance of the proposed approach in terms of AU-PR and G-mean metrics. The results of this experiment are tabulated in Table \ref{win}. From the table it may observed that increasing the window length up to some point typically improves the performance. This may be justified by the fact that incorporating more context from neighbouring data points provides complementary information for classification. However, moving beyond a time length of 100 does not always improves the performance. This behaviour may justified as fine details related to anomalies in the data may be lost when considering long windows. Furthermore, as computation of the signature kernel involves approximating differential operators with finite differences, the longer the time window, the more likely that approximation errors may be accumulated, and hence, adversely affecting the performance. On average, a window length of 100 yields the overall best trade-off between performance and computational complexity across different datasets.

\begin{table}[t!]
\renewcommand{\arraystretch}{1}
\caption{Effect of window size on the performance in the proposed method on multiple data sets in terms of AU-PR (denoted as ``A'') and G-mean (denoted as ``G'') (\%).}
\label{win}
\centering
\begin{adjustbox}{max width=1\columnwidth}
\begin{tabular}{lccccccccccccccc}
\hline
Dataset&\multicolumn{2}{c}{MSL}&\multicolumn{2}{c}{SMAP}&\multicolumn{2}{c}{SMD}&\multicolumn{2}{c}{SWAT}&\multicolumn{2}{c}{WADI}\\
\cmidrule(lr){2-3} \cmidrule(lr){4-5} \cmidrule(lr){6-7}\cmidrule(lr){8-9}\cmidrule(lr){10-11}\\
&A&G&A&G&A&G&A&G&A&G\\
\hline
200 &91.5&67.0 &90.6 &68.3 & 97.3&67.9& 88.9&52.0&93.5&47.2&\\
100 &92.1&66.6&87.1&70.4&97.5&68.5&89.4&52.8&94.9&54.5&\\
50 &90.9 &66.7 &84.5 &64.9 & 97.6& 71.3& 89.7&53.6 &94.3&50.5&\\
20 &90.6&65.9&83.0 & 57.9& 97.8&72.4&92.9&62.6& 93.5& 46.9&\\
\hline
\end{tabular}\end{adjustbox}
\end{table}

\subsubsection{Effect of pseudo-negative training samples}
Finally, in this section, the effect of using pseudo-negative samples generated by the method of \cite{DARBAN2025110874} on the performance is investigated. For this purpose, we train the proposed approach using purely positive samples and compare its performance to the case where pseudo-negative data is used for training.  The results corresponding to this experiment are reported in Table \ref{negimpact}. From the table, one observes that using pseudo-negative samples on all datasets generally improves the performance. This is despite the fact the method of generating pseudo-negative samples employed in \cite{DARBAN2025110874} may not cover all possible anomaly types, yet it appears to be useful in refining the decision boundary in the proposed technique.

\begin{table}[t!]
\renewcommand{\arraystretch}{1}
\caption{The effect of pseudo-negative training samples for time series anomaly detection on multiple data sets in terms of AU-PR (denoted as ``A'') and G-mean (denoted as ``G'') (\%).}
\label{negimpact}
\centering
\begin{adjustbox}{max width=1\columnwidth}
\begin{tabular}{lccccccccccccccc}
\hline
Dataset&\multicolumn{2}{c}{MSL}&\multicolumn{2}{c}{SMAP}&\multicolumn{2}{c}{SMD}&\multicolumn{2}{c}{SWAT}&\multicolumn{2}{c}{WADI}\\
\cmidrule(lr){2-3} \cmidrule(lr){4-5} \cmidrule(lr){6-7}\cmidrule(lr){8-9}\cmidrule(lr){10-11}\cmidrule(rr){12-13}\\
&A&G&A&G&A&G&A&G&A&G&\\
\hline
with pseudo-neg. &92.1&66.6&87.1&70.4&97.5&68.5&89.4&52.8&94.9&54.5\\
without pseudo-neg. &91.7&65.6&86.8&69.6&97.5 &69.7 &89.7 &53.3 &94.1&49.7 \\
\hline
\end{tabular}\end{adjustbox}
\end{table}

\section{Conclusion}
\label{conc}
In this study, we proposed manifold information in the context of one-class learning for time series anomaly detection. Drawing on an existing Representer theorem, we formulated the optimisation problem for the proposed approach in the dual space. By illustrating that the learning problem of the new method corresponds to that of the large-margin $\ell_p$-SVDD approach but with a modified kernel matrix, we presented an optimisation approach for the proposed method. We experimentally evaluated the proposed method on multiple multidimensional time-series datasets to show its superior performance compared with the existing methods. Furthermore, using Rademacher complexities, we theoretically illustrated that the incorporation of a manifold regularisation term improves generalisation performance of the method. An important observation regarding the proposed approach is that the experimental effectiveness of the proposed method for other learning problems relies on one's ability to derive structure from the marginal distribution and on how much that structure uncovers the underlying truth.

\printcredits

\bibliographystyle{cas-model2-names}

\bibliography{cas-refs}



\end{document}